\documentclass{article}
\usepackage[utf8]{inputenc}

\usepackage{graphicx}
\usepackage{subcaption}

\title{A New Backpropagation Algorithm without Gradient Descent}
\author{
  Varun Ranganathan\\
  Student at PES University \\
  \texttt{varunranga1997@hotmail.com}
  \and
  S. Natarajan\\
  Professor at PES University \\
  \texttt{natarajan@pes.edu}
}
\date{January 2018}

\begin{document}

    \maketitle
    
    \begin{abstract}
        The backpropagation algorithm, which had been originally introduced in the 1970s, is the workhorse of learning in neural networks. This backpropagation algorithm makes use of the famous machine learning algorithm known as Gradient Descent, which is a first-order iterative optimization algorithm for finding the minimum of a function. To find a local minimum of a function using gradient descent, one takes steps proportional to the negative of the gradient (or of the approximate gradient) of the function at the current point. In this paper, we develop an alternative to the backpropagation without the use of the Gradient Descent Algorithm, but instead we are going to devise a new algorithm to find the error in the weights and biases of an artificial neuron using Moore-Penrose Pseudo Inverse. The numerical studies and the experiments performed on various datasets are used to verify the working of this alternative algorithm.\newline
        \newline
        Index Terms – Machine Learning, Artificial Neural Network (ANN), Backpropagation, Moore-Penrose Pseudo Inverse.
    \end{abstract}
    
    \section{Introduction}
        Artificial Neural Network (ANN), inspired by biological neural networks, is based on a collection of connected units or nodes called artificial neurons. These systems are used as a learning algorithm which tries to mimic how the brain works. ANNs are consider as universal function approximators, that is, it can approximate the function for the data sent through it. It is based on the multilayer perceptron \cite{Perceptron} model which is a class of feedforward artificial neural networks, consisting of at least three layers of models. Learning occurs in the perceptron by changing connection weights after each piece of data is processed, based on the amount of error in the output compared to the expected result. This is an example of supervised learning, and is carried out through backpropagation, a generalization of the least mean squares algorithm in the linear perceptron. The multilayer perceptron model coupled with the backpropagation algorithm gave rise to the Artificial Neural Network, which can be effectively and efficiently used as a learning algorithm.\newline
        \newline
        Backpropagation \cite{Backpropagation} is a method used in artificial neural networks to calculate the error contribution of each neuron after a batch of data is processed. It is commonly used by the gradient descent optimization algorithm to adjust the weight of neurons by calculating the gradient of the loss function. This technique is also sometimes called backward propagation of errors, because the error is calculated at the output and distributed back through the network layers. It also requires a known, desired output for each input value — it is therefore considered to be a supervised learning method.\newline
        \newline
        Gradient Descent \cite{GradientDescent} is an iterative approach that takes small steps to reach to the local minima of the function. This is used to update the weights and biases of each neuron in a neural network. Gradient descent is based on the observation that if the multivariable function $F(x)$ is defined and differentiable in a neighborhood of a point $a$, then $F(x)$ decreases fastest if one goes from $a$ in the direction of the negative gradient of $F$ at $a$, $- \Delta F(a)$. It follows that, if $a_{n+1} = a_n \, – \, \gamma \Delta F(a_n)$ for $\gamma$ small enough, then $F(a_n) >= F(a_{n+1})$.\newline 
        \newline
        In other words, the term $\gamma \Delta F(a)$ is subtracted from $a$ because we want to move against the gradient, toward the minimum. With this observation in mind, one starts with a guess $x_0$ for a local minimum of $F$, and considers the sequence $x_0$, $x_1$, $x_2$, ... such that $x_{n+1}$ = $x_n \, – \, \gamma _n \Delta F(x_n)$, for $n >= 0$. We have $F(x_0) >= F(x_1) >= F(x_2) >= ...$, so hopefully the sequence $x_n$ converges to the desired local minimum.\newline
        \newline
        Even though this method works well in general, it has a few limitations. Firstly, due to the iterative nature of the algorithm, it takes a lot of time to converge to the local minima of the function. Secondly, gradient descent is relatively slow close to the minimum: technically, its asymptotic rate of convergence is inferior to many other methods. Thirdly, the gradient methods are ill-defined for non-differentiable functions.\newline
        \newline
        During the paper we will be referring to the Moore-Penrose Pseudo Inverse \cite{PseudoInverse}. In mathematics, and in particular linear algebra, a pseudoinverse A+ of a matrix A is a generalization of the inverse matrix. The most widely known type of matrix pseudoinverse is the Moore–Penrose inverse, which was independently described by E. H. Moore in 1920, Arne Bjerhammar in 1951, and Roger Penrose in 1955. A common use of the pseudoinverse is to compute a `best fit' (least squares) solution to a system of linear equations that lacks a unique solution. Another use is to find the minimum (Euclidean) norm solution to a system of linear equations with multiple solutions. The pseudoinverse facilitates the statement and proof of results in linear algebra. The pseudoinverse is defined and unique for all matrices whose entries are real or complex numbers. It can be computed using the singular value decomposition.\newline
        \newline
        In this paper, we formulate another method of finding the errors in weights and biases of the neurons in a neural network. But first, we would like to present a few assumptions made in the model of the neural network, to make our method feasible.
        
    \section{Modifications to the neuron structure}
        We have made one change to the structure of an artificial neuron. We assume that there is a weight and bias associated for each input, that is, each element in the input vector is multiplied by a weight and a bias is added to it. This is a slight alteration from the traditional artificial neuron where there is a common bias applied to the overall output of the neural network. This change will not alter the goal or the end result of a neural network. The proof for this statement is shown below:\newline
        \newline
        For input vector of size `n':
        \begin{equation}
            c_1w_1 + b_1 + c_2w_2 + b_2 + c_3w_3 + b3 ... c_nw_n + b_n
        \end{equation}
        \begin{equation}
            = c_1w_1 + c_2w_2 +c_3w_3 ... c_nw_n + b
        \end{equation}
        Where :
        \newline
        \begin{equation}
            b = b_1 + b_2 + b_3 .. b_n
        \end{equation}
        \newline
        Therefore, having a separate bias for each input element will make no difference to the end result.
        \newline

        \begin{figure}[h!]
            \centering
            \includegraphics[scale=0.33]{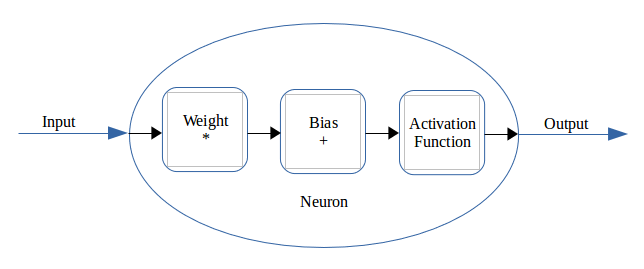}
            \caption{Neuron}
            \label{fig:1D Neuron}
        \end{figure}
        
    \section{The New Backpropagation Algorithm}
        
        \subsection{Calculating new weights and biases for a neuron}
            Taking one neuron at a time, there is one input entering into the neuron, which is multiplied by some weight and a bias is added to this product. This value is then sent through an activation function, and the output from activation function is taken as the output of the neuron.\newline
            \newline
            Let $C$ be the input into the neuron,\newline
            the original weight applied to that input is $w$\newline
            and the original bias applied to that input is $b$.\newline
            \newline
            Let $x$ be the output given initially when the input $C$ passes through the neuron.\newline
            \newline
            Let $x_n$ be the output that we require.\newline
            Based on the required output, we will require a different weight and bias value, say $w_n$ and $b_n$ respectively.\newline
            \newline
            The original output is calculated as,
            \begin{equation}
                Cw + b = x
            \end{equation}
            But, we required $x_n$ as the output.
            Therefore,
            \begin{equation}
                Cw_n + b_n = x_n
            \end{equation}
            \newline
            Let,
            \begin{equation}
                w_n = w - \Delta w
            \end{equation}
            \begin{equation}
                b_n = b - \Delta b
            \end{equation}
            Where,\newline
            $\Delta  w$ is the error in the weight and,\newline
            $\Delta  b$ is the error in the bias.\newline
            \newline
            \begin{equation}
                Cw_n + b_n = x_n
            \end{equation}
            \begin{equation}
                C(w - \Delta w) + (b - \Delta b) = x_n
            \end{equation}
            \begin{equation}
                C(w - \Delta w) + (b - \Delta b) = x_n
            \end{equation}
            \begin{equation}
                (Cw + b) - (C\Delta w + \Delta b) = x_n
            \end{equation}
            \begin{equation}
                x - x_n = (C\Delta w + \Delta b)
            \end{equation}
            Therefore,
            \begin{equation}
                C\Delta w + \Delta b = (x - x_n)
            \end{equation}
            Now,
            \begin{equation}
                [
                \begin{array}{cc}
                    C & 1
                \end{array}
                ]
                \times
                [
                \begin{array}{cc}
                    \Delta w \\ \Delta b
                \end{array}    
                ]
                =
                [
                \begin{array}{cc}
                    (x - x_n)
                \end{array}    
                ]
            \end{equation}
            \newline
            \begin{equation}
                [
                \begin{array}{cc}
                    \Delta w \\ \Delta b
                \end{array}    
                ]
                =
                [
                \begin{array}{cc}
                    C & 1
                \end{array}
                ]^{-1}
                \times
                [
                \begin{array}{cc}
                    (x - x_n)
                \end{array}    
                ]
            \end{equation}
            \newline
            But,
            $[\begin{array}{cc} C & 1\end{array}]$ is not a square matrix.\newline
            Therefore,
            We will have to find the Moore-Penrose Pseudo-Inverse of the matrix $[\begin{array}{cc} C & 1\end{array}]$.\newline
            \begin{equation}
                [
                \begin{array}{cc}
                    \Delta w \\ \Delta b
                \end{array}    
                ]
                =
                [
                \begin{array}{cc}
                    C & 1
                \end{array}
                ]^{+}
                \times
                [
                \begin{array}{cc}
                    (x - x_n)
                \end{array}    
                ]
            \end{equation}
            \newline
            After obtaining $\Delta w$ and $\Delta b$, change the original weight and bias to the new weight and bias in accordance to,
            \begin{equation}
                w_n = w - (\Delta w * \alpha)
            \end{equation}
            \begin{equation}
                b_n = b - (\Delta b * \alpha)        
            \end{equation}
            where $\alpha$ is the learning rate.
            
        \subsection{Tackling multiple inputs}
            The above mentioned method of changing weights and biases of the neuron can be extended for a vector input of length $n$.\newline

            \begin{figure}[h!]
                \centering
                \includegraphics[scale=0.33]{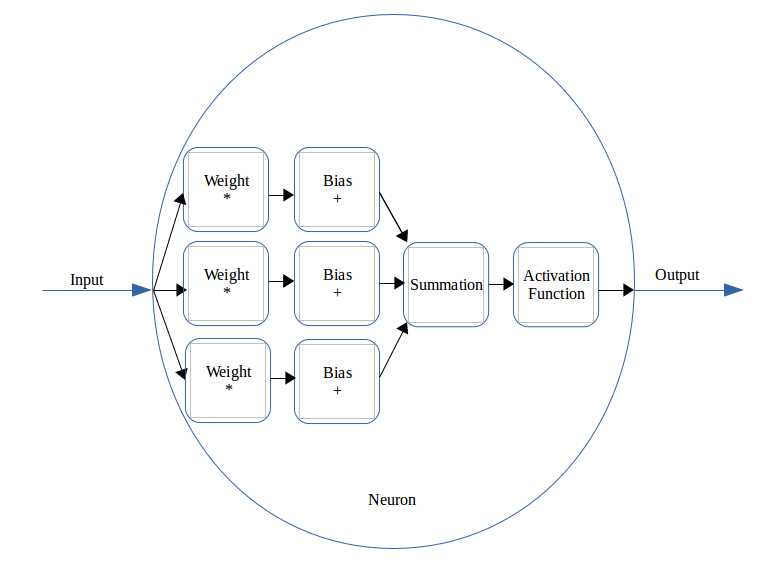}
                \caption{Neuron for vector length of `n'}
                \label{fig:ND Neuron}
            \end{figure}
            
            Let the input vector $C$ belong to the $n^{th}$ dimension.\newline
            \newline
            In this case, each element of the input vector will be multiplied by its respective weight in the neuron, and a bias will be added to each of the products. Therefore, there will be $n$ input elements, $n$ corresponding weights and biases, and $n$ outputs from each weight-bias block. These outputs are added up to give one single output and passed on the activation function.\newline
            
            During the backpropagation stage, the desired output is distributed amongst all the weight-bias pairs, such that, for a block of weight and bias $i$ ($w_i$, $b_i$), the required output for that block will be $1/n$ of the required output.\newline
            \newline
            That is,
            For all weight-bias blocks ($w_i$, $b_i$)
            \begin{equation}
                x_{n_{i}} = x_n / n
            \end{equation}
            \newline
            The weights and biases are initialized to random values in the beginning, that is, absolute weightage given to each element in the input vector is randomized. The relative weightage given to each element in the input vector should be the same. Each weight-bias block will give the same output, so that the cumulative output will give us the required answer. Therefore, this method of dividing the weights will work.\newline
        
        \subsection{Activation Function for non-linearity}
            To achieve non-linearity, the general approach taken is to pass the summation of the output from all weight-bias pairs through a non-linear activation function \cite{SigmoidTanh}. During the backpropagation phase, to correct the weights and biases values of the neuron, we cannot simply pass the actual output vector required. If we do so, it will change the weights and biases as though there is no activation function, and when the forward propagation of the same vector occurs, the neuron outputs will go through the activation function, and give a wrong result. Therefore, we must pass the output vector required through the inverse of the activation function. We need to make sure that we will have to choose an activation function such that its domain and range are the same, so as to avoid math errors and to avoid loss of data. The new vector after applying the inverse activation function is the actual vector sent to the layers of the network during the backpropagation phase.\newline
        
        \subsection{Network Architecture}
        
            \begin{figure}[h!]
                \centering
                \includegraphics[scale=0.33]{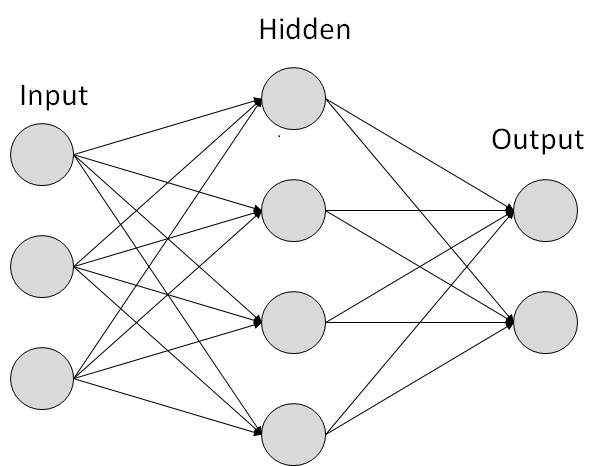}
                \caption{Neural Network Representation}
                \label{fig:Typical Neural Network}
            \end{figure}
            
            Figure 3 shows the representation of a neural network. Each neuron outputs one value. The output of every neuron in one layer is sent as the input to every neuron in the next layer. Therefore, each layer can be associated with a buffer list, so that the output from each neuron in that layer can be stored and passed on to the next layer as input. This would help in the implementation of a neural network by simplifying the forward propagation task.\newline
            
            \begin{figure}[h!]
                \centering
                \includegraphics[scale=0.33]{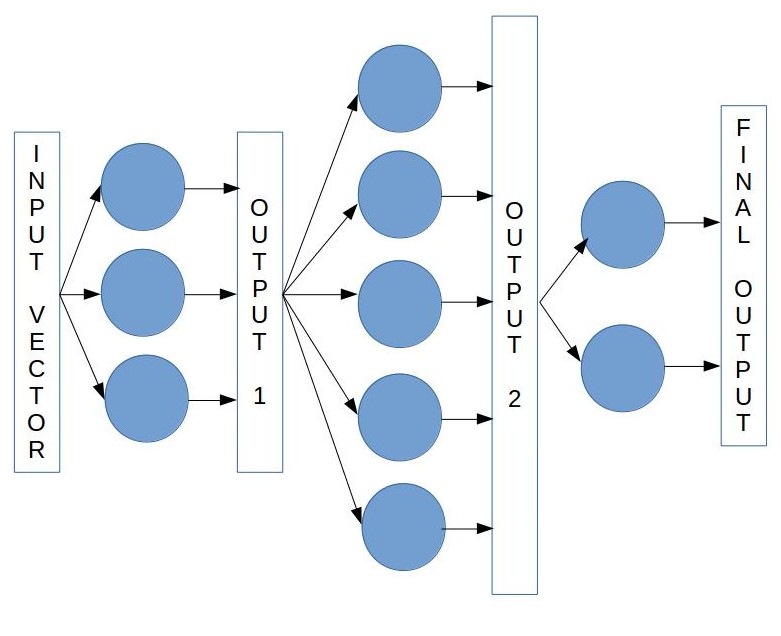}
                \caption{Neural Network Implementation}
                \label{fig:Neural Network Implementation}
            \end{figure}
            
            The input forward propagates through the network and at the last (output) layer it gives out an output vector. Now, for this last layer, the required output is known. Therefore the weights and biases of the neurons of the last layer can be easily changed.\newline
            \newline 
            We do not know the required output vectors for the previous layers. We can only make a calculated guess. Using a simple intuition by asking ourselves the question, "What should be the input (which is the output vector of the previous layer) to the last layer such that the output would be correct?", we can arrive at a conclusion that the input, which would be the correct required output for the previous layer, is a vector which should have given no error in the output of the last layer. This can be illustrated by the following equations.\newline
            If
            \begin{equation}
                C*w + b = x  
            \end{equation}
            Then what $C_n$ vector will satisfy the equation $C_n*w + b = x$
            \begin{equation}
                C_n = (x_n - b) / w        
            \end{equation}
            \newline
            This approach can be extended to all the previous layers.\newline
            \newline
            Another issue arises that many neurons will give their own ‘required’ input, so that their outputs will be correct. This could happen in a multiclass classification problem, wherein the output vector required is one-hot encoded vector (where the element of the vector at the position of the required class is 1, and the other elements in the vector are 0). Therefore, we take the average of all vectors. This will give an equal weightage of all the feedbacks from each neuron. Pass this averaged required input vector to the previous layers as the required output from that layer.\newline
            \newline
            
        This concludes the complete working of the neural network with our devised backpropagation algorithm.
        
    \section{Differences with Extreme Learning Machines}
        Extreme learning machines \cite{ExtremeLearningMachines} are feedforward neural network for classification, regression, clustering, sparse approximation, compression and feature learning with a single layer or multilayers of hidden nodes, where the parameters of hidden nodes (not just the weights connecting inputs to hidden nodes) need not be tuned. These hidden nodes can be randomly assigned and never updated (i.e. they are random projection but with nonlinear transforms), or can be inherited from their ancestors without being changed. In most cases, the output weights of hidden nodes are usually learned in a single step, which essentially amounts to learning a linear model.\newline
        \newline
        Even though both the method use the Moore-Penrose Pseudo Inverse, there are a few significant differences between the ELM and the proposed backpropagtion method explained in this paper. The ELM is a feedforward network which is aims at replacing the traditional artificial neural network, whereas this paper provides an alternative for the backpropagation algorithm used in traditional artificial neural networks. The ELM algorithm provides only a forward propagation technique to change the weights and bias of the neurons in the last hidden layer, whereas we have provided a method of backpropagation to change the weights and biases of all neurons in every layer.
    
    \section{Results}
        
        \subsection{Telling-Two-Spirals-Apart Problem}
            Alexis P. Wieland proposed a useful benchmark task for neural networks: distinguishing between two intertwined spirals. Although this task is easy to visualize, it is hard for a network to learn due to its extreme non-linearity. In this report we exhibit a network architecture that facilitates the learning of the spiral task, and then compare the learning speed of several variants of the backpropagation algorithm.\newline
            \newline
            In our experiment, we are using the spiral dataset which contains 193 data points of each class. We have decided to model the network with a 16-32-64-32-2 configuration, with `Softplus' activation function on all neurons of the network. We trained the model for 1000 epochs, with a learning rate of 0.0002.\newline
            
            \begin{figure}[h!]
                \centering
                \includegraphics[scale=0.5]{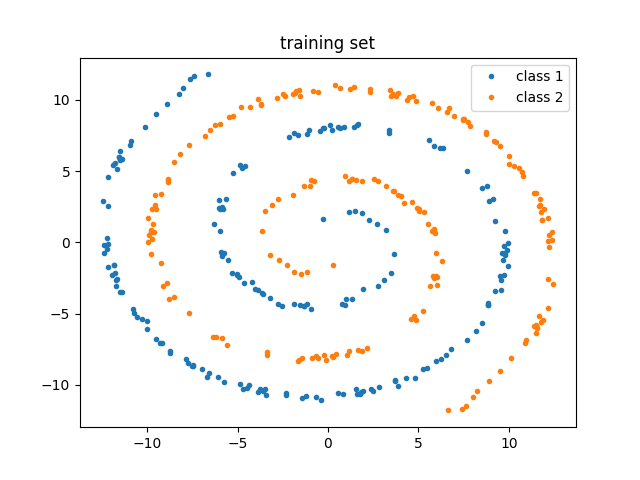}
                \caption{Training data points for the Two-Spirals problem}
                \label{fig:training_data_points}
            \end{figure}
            
            \begin{figure}[h!]
                \centering
                \includegraphics[scale=0.5]{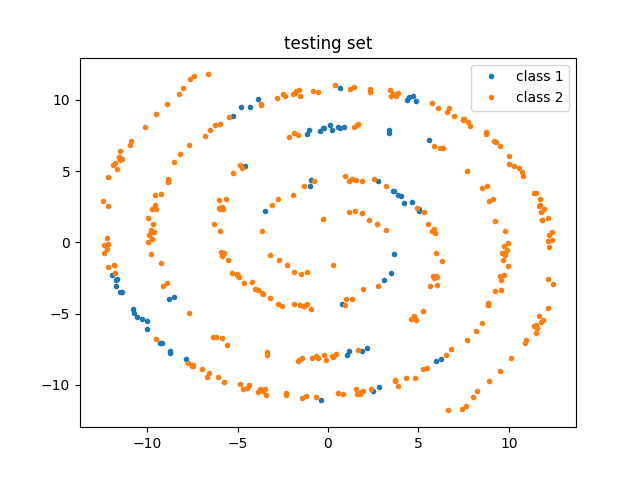}
                \caption{Testing data points for the Two-Spirals problem}
                \label{fig:testing_data_points}
            \end{figure}
            
            From the above 2 figures, we can see that although it doesn't distinguish between the two spirals very well, we are able to get an accuracy of about 63\%. This is due to the fact that the Softplus activation function is not the recommended activation function for this particular problem. The recommended activation function is `Tanh' but, due to the fact that the domain of inverse of the Tanh function lies between $(-1,1)$ and not between $(-\infty,+\infty)$, it cannot be used in our backpropagation method without causing some loss of data.\newline
            \newline
            Looking at figure 6, we can observe the non-linearity in the classification of the two sets of spirals, which proves that this backpropagation method is working.
        
        \subsection{Separating-Concentric-Circles Problem}
            Another type of natural patterns is concentric rings. As a test, we use the sklearn.dataset.make\_circles function to create 2 concentric circles with each 100 data points, which were respectively assigned to two classes. We used an artificial neural network model with the configurations 16-64-32-2, again using the `Softplus' activation function on all neurons of the network. We trained the model for 1000 epochs with a learning rate of 0.00001.\newline
            
            \begin{figure}[h!]
                \centering
                \includegraphics[scale=0.5]{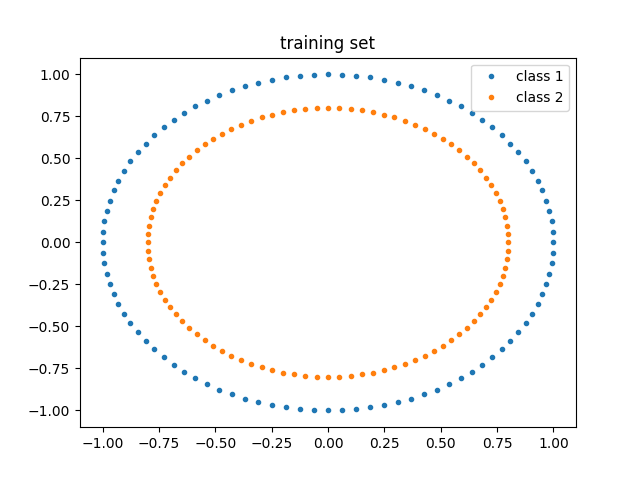}
                \caption{Training data points for the Concentric-Circles problem}
                \label{fig:training_data_points}
            \end{figure}
            \begin{figure}[h!]
                \centering
                \includegraphics[scale=0.5]{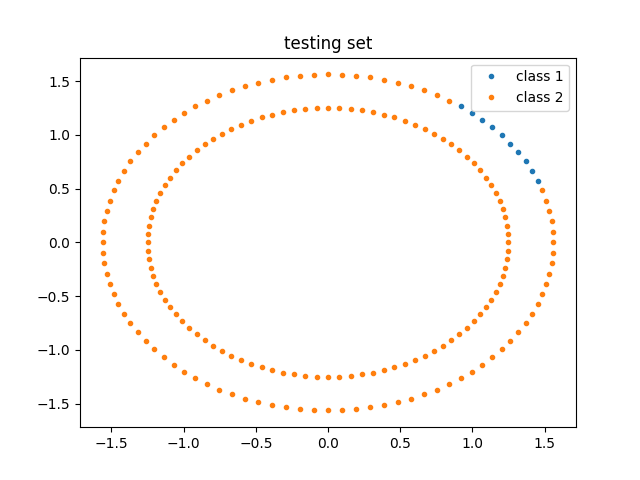}
                \caption{Testing data points for the Concentric-Circles problem}
                \label{fig:testing_data_points}
            \end{figure}

            Observing figure 8, we can see that there is a slight non-linearity in the classification of the 2 points. We can observe an accuracy rate of 61\%. This low accuracy can again be justified with the fact that the softplus activation function is not suitable for such types of data.
            
        \subsection{XOR Problem}
            Continuing our tests on this alternate algorithm, we create a dataset with 1000 data points with each data sample containing 2 numbers, and 1 class number. If the 2 numbers are both positive or negative, the class is 0, else, the class number is 1. The XOR function is applied on the sign of the number.\newline
            \newline
            Our model was of configuration 4-8-16-32-1 where `Softplus' activation function is applied by all neurons. The learning rate was set to 0.0001 and the network was trained for 100 epochs.\newline
            
            \begin{figure}[h!]
                \centering
                \includegraphics[scale=0.5]{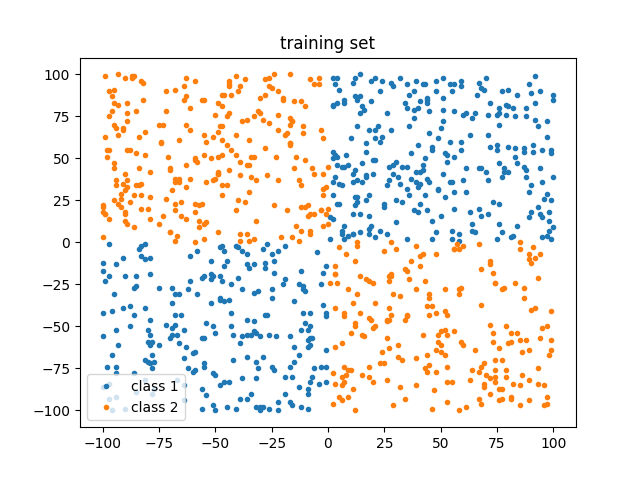}
                \caption{Training data points for the XOR problem}
                \label{fig:training_data_points}
            \end{figure}
            \begin{figure}[h!]
                \centering
                \includegraphics[scale=0.5]{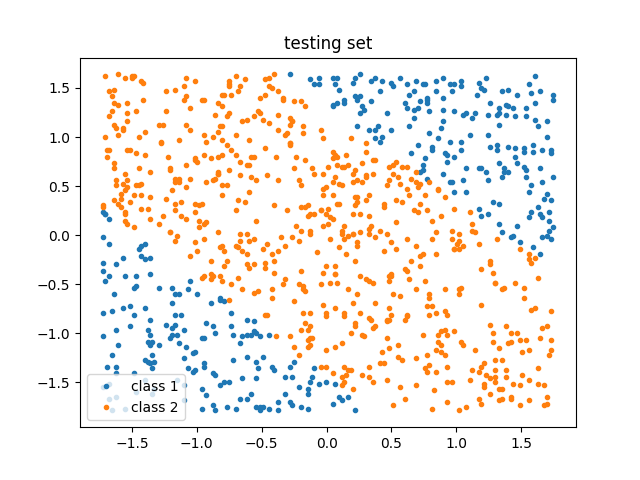}
                \caption{Testing data points for the XOR problem}
                \label{fig:testing_data_points}
            \end{figure}
            
        A validation accuracy of 81\% was achieved.
            
        \subsection{Wisconsin Breast Cancer Dataset}
            To further test our neural network model, we used a real-world dataset in testing our neural network. This dataset contains 699 samples, where each sample has 10 attributes as the features, and 1 class attribute. This dataset is taken from the UCI Machine Learning Repository, where samples arrive periodically as Dr. Wolberg reports his clinical cases.\newline
            \newline
            The model had a configuration of 16-2, and the `Softplus' activation function is applied by all neurons. We trained the model for 1000 epochs with a learning rate of 0.0001. We could observe that the validation accuracy reached upto 90.4\% at the 78th epoch. Even though the values of validation error and training error are erratic in the start, they seem to reach an almost constant value after some number of epochs.\newline
            
            \begin{figure}[h!]
                \centering
                \includegraphics[scale=0.5]{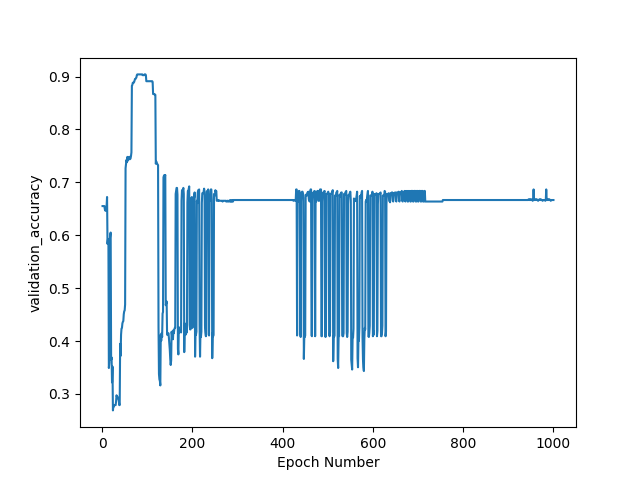}
                \caption{Validation Accuracy while training for Wisconsin Breast Cancer Dataset}
                \label{fig:training_data_points}
            \end{figure}
            \begin{figure}[h!]
                \centering
                \includegraphics[scale=0.5]{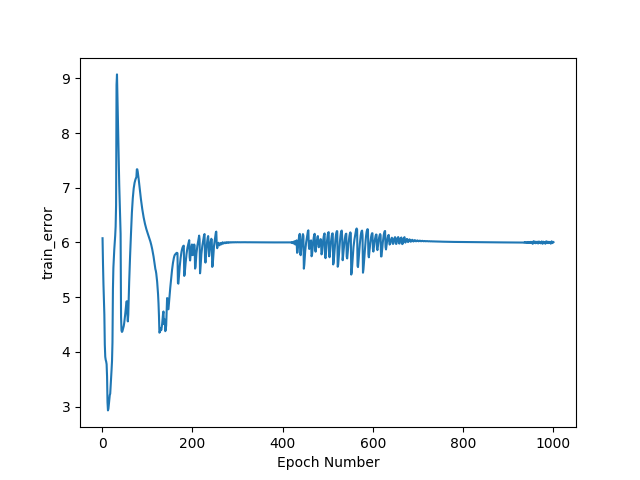}
                \caption{Training Error while training for Wisconsin Breast Cancer Dataset}
                \label{fig:testing_data_points}
            \end{figure}
            \begin{figure}[h!]
                \centering
                \includegraphics[scale=0.5]{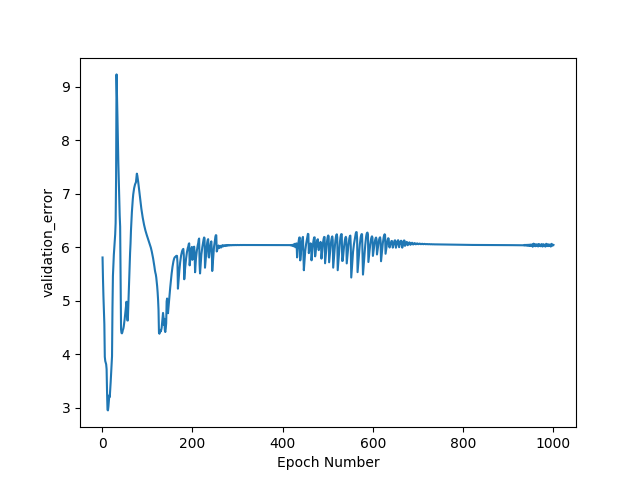}
                \caption{Validation Error while training for Wisconsin Breast Cancer Dataset}
                \label{fig:testing_data_points}
            \end{figure}

        From the above experiments, we can conclude that the Softplus activation function is more suited to the Wisconsin Breast Cancer Dataset and that our proposed backpropagation algorithm truly works.
            
    \section{Discussions and Conclusion}
        From the above stated facts and results, we can observe a few properties with this method. This proposed method of backpropagation can be used very well with activation functions where the domain of the activation function matches the range of its inverse. This property eases the requirement that the activation function must be differentiable. Therefore, ReLU-like activation functions such as LeakyReLU, Softplus, S-shaped rectified linear activation unit (SReLU), etc. will be a good match with this method.\newline
        \newline
        Further optimizations must be made to this method, so that, it can be efficiently used. The requirement of a different type of activation function could accelerate the discovery of many more activation functions which could fit various different models.\newline 
        \newline
        We believe that because this backpropagation method suits ReLU-like \cite{ReLU} activation functions, it can be enhanced to be used in the fields of biomedical engineering, due to the asymmetric behaviour of data collected in such fields where the number of data points in different classes are not balanced. Possibly in the future, if a suitable replacement for activation functions, such as Sigmoid and Tanh, are created, this method could be used more frequently.

\end{document}